\documentclass[10pt,twocolumn,letterpaper]{article}

\usepackage{cvpr}              

\usepackage{graphicx}
\usepackage{amsmath}
\usepackage{amssymb}
\usepackage{booktabs}
\usepackage{bm}

\usepackage[accsupp]{axessibility}
\usepackage[pagebackref=true,breaklinks=true,colorlinks,bookmarks=false]{hyperref}
\pdfoutput=1
\begin{document}

\title{Modeling Image Composition for Complex Scene Generation}


\newcommand*\samethanks[1][\value{footnote}]{\footnotemark[#1]}
\author{
Zuopeng Yang$^{1*}$ ~~~~~ Daqing Liu$^{2*}$ ~~~~~ Chaoyue Wang$^{3}$ ~~~~~ Jie Yang$^{1\dag}$ ~~~~~ Dacheng Tao$^{2,3}$ \\[1.2mm]
$^{1}$Shanghai JiaoTong University ~~~~~~~ $^{2}$JD Explore Academy, JD.com ~~~~~~~ $^{3}$ The University of Sydney\\
\tt\small \{yzpeng, jieyang\}@sjtu.edu.cn, chaoyue.wang@outlook.com, \{liudq.ustc, dacheng.tao\}@gmail.com
}

\twocolumn[{%
\maketitle%
\vspace{-2.5em}
\begin{center}
    \includegraphics[width=1.00\textwidth, trim=0em 0em 0em 0em, clip]{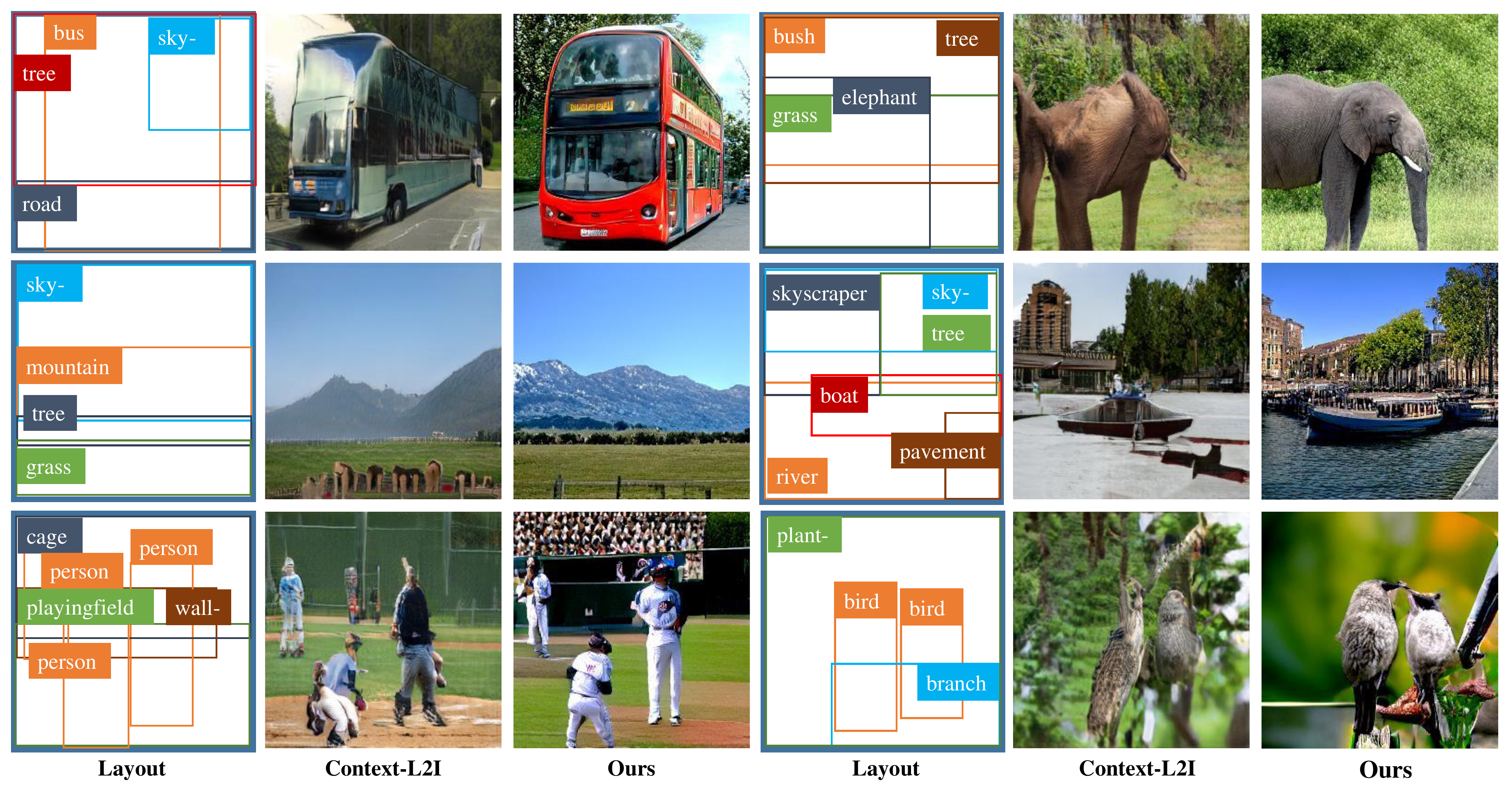}
  \vspace{-20px}
  \captionof{figure}{By specifying the layouts, we compared several results generated by recent CNN-based method \cite{he2021context} and our proposed TwFA. Our approach enables transformers to synthesize high-quality images containing \textbf{multiple objects with complex structures} from layouts (bounding boxes with categories).}
  \label{fig:banner}
\end{center}
}]

\let\thefootnote\relax\footnote{$^{*}$Equal Contribution. $^{\dag}$Corresponding author.}
\let\thefootnote\relax\footnote{This research is partly supported by NSFC, China (No: 61876107, U1803261), and Dr. Chaoyue Wang is supported by ARC FL-170100117.}

\begin{abstract}
We present a method that achieves state-of-the-art results on challenging (few-shot) layout-to-image generation tasks by accurately modeling textures, structures and relationships contained in a complex scene. After compressing RGB images into patch tokens, we propose the Transformer with Focal Attention (TwFA) for exploring dependencies of object-to-object, object-to-patch and patch-to-patch. Compared to existing CNN-based and Transformer-based generation models that entangled modeling on pixel-level\&patch-level and object-level\&patch-level respectively, the proposed focal attention predicts the current patch token by only focusing on its highly-related tokens that specified by the spatial layout, thereby achieving disambiguation during training. Furthermore, the proposed TwFA largely increases the data efficiency during training, therefore we propose the first few-shot complex scene generation strategy based on the well-trained TwFA. Comprehensive experiments show the superiority of our method, which significantly increases both quantitative metrics and qualitative visual realism  with respect to state-of-the-art CNN-based and transformer-based methods. Code is available at \url{https://github.com/JohnDreamer/TwFA}.
\end{abstract}

\vspace{-1em}

\section{Introduction}
Generating photo-realistic images is the ever-lasting goal in computer vision.
Despite achieving remarkable progress on image generation for both simple scenario, \textit{e.g.}, faces, cars, and cats~\cite{karras2019style,richardson2021encoding,karras2020analyzing}, and single object, \textit{e.g.}, ImageNet~\cite{brock2018large,zhang2019self}, the image generation for complex scenes composed of multiple objects of various categories is still a challenging problem.

In this paper, we focus on one representative complex scene image generation task, layout to image generation~\cite{zhao2019image} (L2I), which aims to generate complex scenes conditioned on specified layouts.
The layout, as illustrated in Figure~\ref{fig:banner}, consists of a set of object bounding boxes and corresponding categories, thus providing a sketch of the expected complex scene image.
Compared with other conditions for complex scene generation, including textual descriptions~\cite{reed2016generative}, scene graphs~\cite{johnson2018image,lin2020gps}, and segmentation masks~\cite{li2021image}, layouts are much more user-friendly, controllable and flexible~\cite{zhao2019image}.
Ambitiously, we further propose a new few-shot layout to image generation task (few-shot L2I), which aims to generate complex scenes with a novel object category after providing only a few images containing the novel objects.

As to the complex scene generation, including (few-shot) L2I tasks, the core challenge is how to synthesize a photo-realistic image with reasonable object-level relationships, clear patch-level instance structures, and refined pixel-level textures.
Existing attempts to the L2I task can be divided into two categories, \textit{i.e.}, CNN-based~\cite{zhao2020layout2image,sun2021learning,sylvain2021object,ma2020attribute,li2020bachgan,li2021image} and Transformer-based~\cite{jahn2021high}, according to their generator.
The CNN-based methods deploy an encoder-decoder generator~\cite{johnson2018image,ma2020attribute} where the encoder transfers the layout into an image feature map, and the decoder upsamples the feature map into the target image.
Those methods capture the object relationships in the encoder by a self-attention~\cite{vaswani2017attention} or a convLSTM~\cite{zhao2020layout2image}, and model the instance structures and textures simultaneously in the decoder by upsampling convolutions.
In contrast, the Transformer-based methods tokenize the layout into object tokens and employ a pre-trained compression model to quantize the image into a sequence of discrete patch tokens, thus simplifying the image generation task as an image patch composition task implemented by a Transformer.
Those methods produce the detailed textures with the compression model, and model both relationships and structures by the Transformer.

However, the entangled modeling on patch-level and pixel-level (CNN-based methods) or object-level and patch-level (Transformer-based methods) prevents the model from capturing inherent instance structures, leading to blurry or crumpled objects, and increases the burden on the few-shot learning because the model must learn the two levels information simultaneously with only a few images.
To this end, upon the Transformer-based methods, we propose a Transformer with Focal Attention (TwFA) to separately model image compositions on object-level and patch-level by distinguishing between object and patch tokens.
Different from vanilla self-attention, which neglects the composition prior of spatial layouts, our focal attention further constrains each token can only attend on its related tokens according to the spatial layouts.
Specifically, to model object relationships, an object token attends on all object tokens to capture the global information. To model instance structures, a patch token attends on the object it belongs to and the patches inner the object bounding box.
By the proposed Focal Attention, the TwFA focuses on generating the current patch without any disturbance from other objects or patches thus increasing the data efficiency during training.
Therefore, the focal attention makes the TwFA can fast learn the novel object category with only a few images.

We validate the effectiveness of the proposed TwFA on COCO-stuff~\cite{caesar2018coco,lin2014microsoft} and Visual Genome~\cite{krishna2017visual} datasets.
TwFA improves the state-of-the-arts~\cite{jahn2021high,li2021image} FID score from 29.56 to 22.15 (-25.1\%) on COCO-stuff, and from 19.14 to 17.74 (-7.3\%) on Visual Genome.
Morevoer, TwFA demonstrates the superiority on the few-shot L2I task with strong performance and impressive visualizations.

\section{Related Work}

\noindent \textbf{CNN-based image generation.}
In recent years, a number of CNN-based generative models have been proposed, and achieved significant progress on (un)conditional image generation tasks.
Till now, CNN-based generative models (\textit{e.g.}, GANs~\cite{goodfellow2014generative,8627945,ijcai2017-404}, VAEs~\cite{kingma2013auto}) are good at synthesizing high-resolution and high-fidelity object images, which include but are not limited to flowers, human/animal faces, and buildings~\cite{brock2018large,karras2019style,choi2018stargan}. However, generating complex real-world scenes which include multiple instances with variant layout and scale has still been a challenging task~\cite{sun2019image,li2021image,he2021context}.
To ease the difficulty of synthesizing complex scenes, previous works usually break the tasks into several steps. For example, Layout2Im~\cite{zhao2019image,zhao2020layout2image} models this task as object generation then image generation pipeline, each object is controlled by a certain category code and an uncertain appearance code. For better controllability, LostGANs~\cite{sun2019image,sun2021learning} first synthesize the semantic masks from layouts, and then ISLA-Norm is proposed for generating color images from specific masks and style codes. In addition, some works focus on improving models' generative performance by introducing pseudo supervisions~\cite{li2020bachgan,sylvain2021object}, additional annotations~\cite{ma2020attribute} or fine-grained control~\cite{frolov2021attrlostgan}.
Although CNN-based layout-to-image generation methods have achieved promising performance on texture synthesis, they may still suffer from accurately modeling the dependencies among pixels (or object parts), which hinders the model generated more realistic scene images. 

\noindent \textbf{Transformer-based image generation.}
Recently, transformers not only demonstrate promising results in computer vision tasks~\cite{xu2021vitae,zhang2022vitaev2}, but also show potential on conditional visual content generation \cite{ramesh2021zero,esser2021taming,lin2021m6}. First, a Vector Quantised Generative Adversarial Network (VQ-GAN)~\cite{esser2021taming} is trained to compress images into finite discrete representations/tokens. Then, an autoregressive transformer is trained to model the dependencies between discrete image tokens. Through modeling together with conditional signals, such as text, class labels and keypoints, transformers demonstrated the strong capability of generating semantic controllable images~\cite{esser2021taming}.
For the complex scene generation, \cite{jahn2021high} made a great attempt on synthesizing the high-resolution image from a given layout. Although promising results have been achieved, synthesizing complex scenes that consist of multiple instances and staff is still a challenge task. Since autoregressive transformers have challenges in handling spatial positions~\cite{wu2021rethinking}, they may not accurately model object-object and object-patch relationships. In this paper, the proposed transformer with focal attention can better model the composition of the complex scenes, and leads to better generation performance.

\noindent \textbf{Few-shot image generation.}
Few-shot learning is first explored in discriminative tasks. 
Given limited data from a target domain, neural networks have to overcome training/fine-tuning difficulties, and generalize the pretrained model to target domain~\cite{wang2020generalizing,he2020piecewise,he2019control}. In few-shot image generation, the generative model is trained to synthesize diverse images from the target domain. Previous few-shot image generation methods mainly focused on generating simple patterns and low resolution results~\cite{rezende2016one,reed2018few}. Recently, \cite{wang2020minegan,gu2021lofgan,ojha2021few,li2020few} extend the few-shot generation to objects with similar structures, such as human faces, buildings and cars. However, it has great challenges in performing few-shot generation on complex scenes or novel classes in complex scenes. To our best knowledge, this paper is the first attempt on few-shot complex scene generation. 

\section{Approach}

\begin{figure}[!t]
	\centering
	\includegraphics[width=0.9\linewidth]{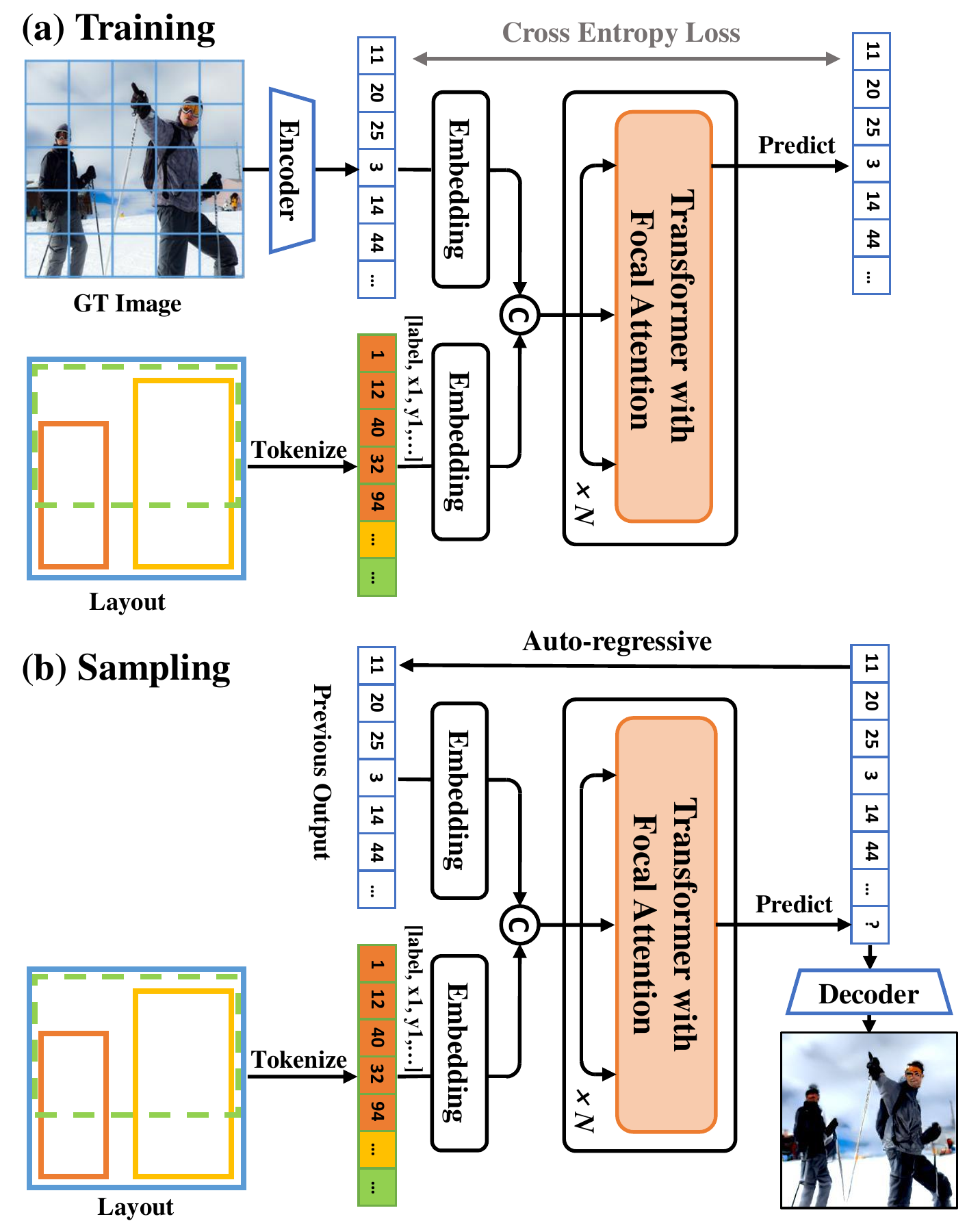}
	\vspace{-2mm}
	\caption{The overview of the proposed Transformer with Focal Attention (TwFA) framework for the L2I task. Given a layout and an image as input, we first 1) tokenize them into sequential discrete object/patch tokens by the embedding/encoder, next 2) predict the next patch token by the TwFA, and then 3) during inference, generate the RGB image from all predicted patch tokens by the decoder.}
	\label{fig:framework}
\end{figure}


\subsection{The Framework}
\label{sec:3.1}

\begin{figure*}[!t]
	\centering
	\includegraphics[width=0.95\linewidth]{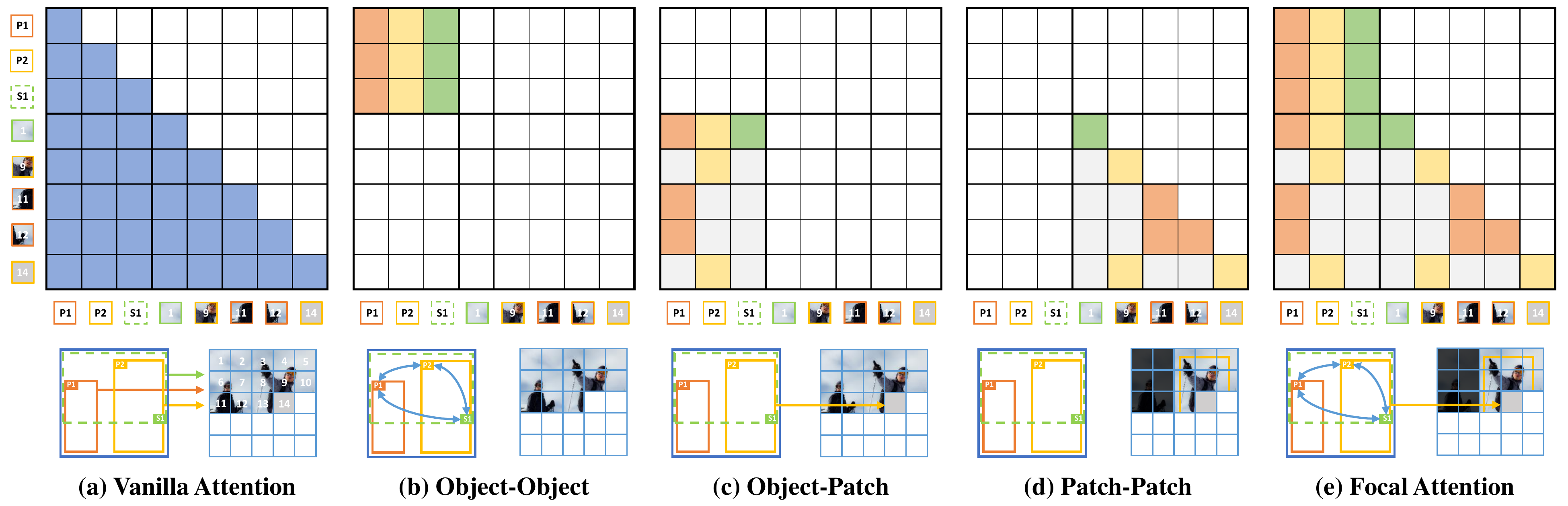}
 	\vspace{-2mm}
	\caption{The illustration of different attention mechanisms with connectivity matrix. (a) Vanilla attention follows a casual mask, neglecting different interactions, (b) Object-object interaction enhances the modeling of object-level relationships; (c) Object-patch interaction makes patches to realize the object categories, (d) Patch-patch interaction introduces the composition prior of layouts, (e) Therefore our focal attention captures both object-level relationships and patch-level structures.
	Colors(orange, yellow, green) correspond to different objects.
	}\vspace{-1 mm}
	\label{fig:attention}
\end{figure*}

As illustrated in Figure~\ref{fig:framework}, the proposed Transformer with Focal Attention (TwFA) for the L2I task generally follows the pipeline of tokenization $\rightarrow$ composition $\rightarrow$ generation, \textit{i.e.}, firstly tokenizes the layout/image into sequential discrete object/patch tokens, secondly predicts the distribution of patch tokens in an auto-regressive manner and composes into a discrete patch token sequence, and finally generates the synthesized image from the patch tokens.

\noindent \textbf{Tokenization.}
Given a layout consists of a set of objects with their bounding boxes and category classes, we directly tokenize it into a sequential object tokens $\bm{c}=\{(l_i, \bm{b}_i)_{i=1}^N$\} with $N$ objects, where $l_i$ denotes the $i^{th}$ object's category, and $b_i=[x_{1i}, y_{1i}, x_{2i}, y_{2i}]$ represents its top-left and bottom right corner positions.

Given an image $\bm{x} \in \mathbb{R}^{3 \times H \times W}$, we tokenize it with an encoder of Vector Quantised Generative Adversarial Network (VQ-GAN~\cite{esser2021taming}) that compresses high-dimensional data into a discretized latent space and reconstructs it.
Specifically, the VQ-GAN encoder $Enc$ tokenizes the image $\bm{x}$ into a collection of indices $\bm{s}$ of codebook entries:
\begin{equation}
    \bm{s} =\{s_1, s_2, \dots, s_M\} = Enc(\bm{x}).
\end{equation}

Here, the codebook actually is a ``vocabulary'' of learned representations $C=\{\bm{e}_i\}_{i=1}^K$ where the vocabulary size is $K$.
The VQ-GAN decoder $Dec$ then tries to reconstruct the original image from these latent codes.
In our method, we use a pretrained generic VQ-GAN~\cite{esser2021taming} and keep the weights frozen without any fine-tuning in our experiments.

\noindent \textbf{Composition.}
After tokenizing the input as layout and patch tokens, we simplify the image generation task into an image composition task, where we can only focus on how to produce the final sequential discrete patch tokens with the proposed Transformer with Focal Attention (TwFA).
In detail, given the object tokens $\bm{c}$ and the generated (or ground-truth) patch tokens $\bm{s}_{<i}$, the TwFA is introduced to model the long-range dependency and predict the probability of the next patch token $s_i$:
\begin{equation}
    p(s_i|\bm{s}_{<i}, \bm{c}) = \mathrm{TwFA}(\bm{s}_{<i}, \bm{c}).
\end{equation}
In an auto-regressive manner, TwFA generates the final patch tokens $\bm{s}$ step-by-step.

\noindent \textbf{Generation.}
With the generated patch tokens $\bm{s}$, we further reconstruct it into a real image by VQ-GAN decoder, as:
\begin{equation}
    \hat{x} = Dec(\bm{s}).
\end{equation}
Since arbitrary appearances of different objects can be encoded into the codebook, VQ-GAN is a useful tool for modeling the texture information.

\noindent \textbf{Training and Sampling.}
To train the TwFA, we directly employ a cross-entropy loss for the sequence prediction task:
\begin{equation}
    \mathcal{L}=-\sum_{i=1}^{M}\mathrm{log}(p(s_i|\bm{s}_{<i}, \bm{c})),
\end{equation}
where $M$ is the token length of images, $s_i$ and $\bm{s}_{<i}$ are tokens tokenized from ground-truth images.
During inference, the ground-truth image and its patch tokens are not available. We leverage multinomial resampling strategy to generate diverse images for the same layout.

\subsection{The Attention}
\label{sec:3.2}
In this part, we first revisit the vanilla attention of Transformer~\cite{vaswani2017attention}, and then elaborate the proposed Focal Attention as illustrated in Figure~\ref{fig:attention}.

\subsubsection{Vanilla Attention}
Given the tokenized object tokens $\bm{c}$ and patch tokens $\bm{s}$, we embed them into the feature $\bm{F}=[\mathrm{Emb}_c(\bm{c}); \mathrm{Emb}_s(\bm{s})]$, where $\mathrm{Emb}_{c/s}$ are two embedding layers, and $[;]$ denotes the concatenate operation. Then the feature $\bm{F}$ is fed into a multi-layer decoding Transformer, which adopts attention mechanism with Query-Key-Value (QKV) model.

Given the queries $\bm{Q}=\bm{W}_Q \bm{F}$, keys $\bm{K}=\bm{W}_K \bm{F}$, and values $\bm{V}=\bm{W}_V \bm{F}$, the vanilla attention is given by:
\begin{equation}
\mathrm{Attention}(\bm{Q}, \bm{K}, \bm{V}) = \mathrm{softmax}\left(\frac{\bm{Q}\bm{K}^{T}}{\sqrt{D_k}} \circ \bm{M}\right) \bm{V},
\end{equation}
where $\bm{M}$ is called connectivity matrix, and $\circ$ denotes element-wise product.
In the standard self-attention mechanism, every token needs to attend to all other generated tokens, \textit{i.e.}, $\bm{M}$ is a causal mask as shown in Figure~\ref{fig:attention} (a), given by:
\begin{equation}
    \bm{M}[i,j] =
    \begin{cases}
    1, & \text{if } j \leq i,\\
    -\infty, & \text{else}.
\end{cases}.
\end{equation}

However, the vanilla attention neglects the different type tokens, \textit{i.e.}, object tokens and patch tokens, in our L2I task, hindering the model to well capture the object-level relationships and patch-level instance structures.
For example, while generating the 14th patch in Figure~\ref{fig:attention} (a), the token attends to all object tokens, including P1, P2, and S1, even though the 14th patch doesn't belong to P1 and S1. Meanwhile, the token also attends to all generated patch tokens, even though not all patches are related to the 14th patch.

\subsubsection{Focal Attention}
To address the above issues in vanilla attention, We carefully design the connectivity matrix to guide the transformer to focus on the related tokens.
To better demonstrate mechanism of our focal attention, we further decompose the connectivity matrix $\bm{M}$ into three areas, \textit{i.e.}, object-object, object-patch, and patch-patch interaction.

\noindent \textbf{Object-Object Interaction.}
To model the object-level relationships and learn the global context for each object, we design the object-object interaction as shown in Figure~\ref{fig:attention} (b).
The dense interaction makes each object can interact with each other to capture the relationships by the multi-layer transformer, which is essential for object structure reasoning, for example, to generate a man kicking a soccer, the human's action can be predicted by the relative position between him and the ball.
While vanilla attention models the context in a single direction, in other words, the first object never know others.
Hence, the connectivity matrix $\bm{M}_{oo}$ in the object-object area can be written as:
\begin{equation}
        \bm{M}_{oo}[i,j] = 1.
\end{equation}

\noindent \textbf{Object-Patch Interaction.}
To make the patch aware of the object category it belongs to, we design the object-patch interaction as shown in Figure~\ref{fig:attention} (c).
We stipulate that a patch of an object attends only on the corresponding object token to enhance the representation of the class, and a patch of a stuff or background can attend on all object tokens to make sure the image surroundings is consistent with the complex scene.
Formally, the connectivity matrix $\bm{M}_{op}$ in the object-patch area is:
\begin{equation}
    \bm{M}_{op}[i,j] = 
    \begin{cases}
    1, & \text{if } \bm{s}_i \text{ relates to } \bm{c}_j, \\
    -\infty, & \text{else}.
\end{cases}
\end{equation}
Here, we explicitly distinguish the instance objects, \textit{e.g.}, man and bus, and stuff objects, \textit{e.g.}, sky and grass.
If $\bm{s}_i$ locates in an instance, we define $\bm{s}_i$ only relates to this one instance.
If $\bm{s}_i$ locates in a stuff, we define $\bm{s}_i$ relates to every objects for we hypothesize the stuff area generation relies on the global information.

\noindent \textbf{Patch-Patch Interaction.}
To ensure the generative consistency in the local area, the relationships between patches need considerations.
As shown in Figure~\ref{fig:attention} (d), the patch-patch interaction realizes the isolation between the instance and background and meanwhile keep the local consistency.
Formally, the connectivity matrix $\bm{M}_{pp}$ in the patch-patch area is:
\begin{equation}
    \bm{M}_{pp}[i,j] = 
    \begin{cases}
    1, & \text{if } \bm{s}_i \text{ and } \bm{s}_j \text{ are neighbors}, j \leq i,\\
    -\infty, & \text{else}.
\end{cases}
\end{equation}
Thanks to the composition prior from the layout, here we define two patches are neighbors if they belong to the same object, or they are both belong to stuffs or background.

By combining the above three types of interaction mechanisms, the connectivity matrix $\bm{M}$ of focal attention can be written as:
\begin{equation}
\bm{M} = \left[
 \begin{matrix}
   \bm{M}_{oo} & -\infty \\
   \bm{M}_{op} & \bm{M}_{pp}
  \end{matrix}
  \right].
\end{equation}
Finally, by unambiguously dealing with the interactions of different types, we increase the data efficiency and make it possible for the complex scene few-shot learning.

\subsection{The Few-Shot L2I}
\label{sec:3.3}
Upon the well-trained TwFA, we are ambitious on few-shot layout to image generation.
Given a novel object class and a few images containing this novel object, we aim at learning a model which can synthesize the complex scene image containing the novel class, while keeping the performance for the base ones.

The few-shot framework is based on the TwFA, we 1) append a new instance embedding to the input layer for the novel class, 2) split the last transformer layer into two, where the first one is for the base classes, and the second one is for the novel class, and 3) insert a token fusion module, which fuse the tokens from two last transformer layer according to the spatial layouts, to produce the final sequential patch tokens.

To train the few-shot framework, we initialize the new instance embedding with the well-trained embedding of its superclass, the second transformer layer with the first one, and fine-tune the new instance embedding and the second last transformer layer while keeping the parameters of pre-trained TwFA frozen.
It's worth noting that thanks to the separate modeling on the hierarchical object-patch-pixel level and the sparse focal attention, our TwFA can fast adopts to new class with only a few images and achieves impressive performance.

\begin{figure*}[!t]
	\centering 
	\includegraphics[width=0.95\linewidth]{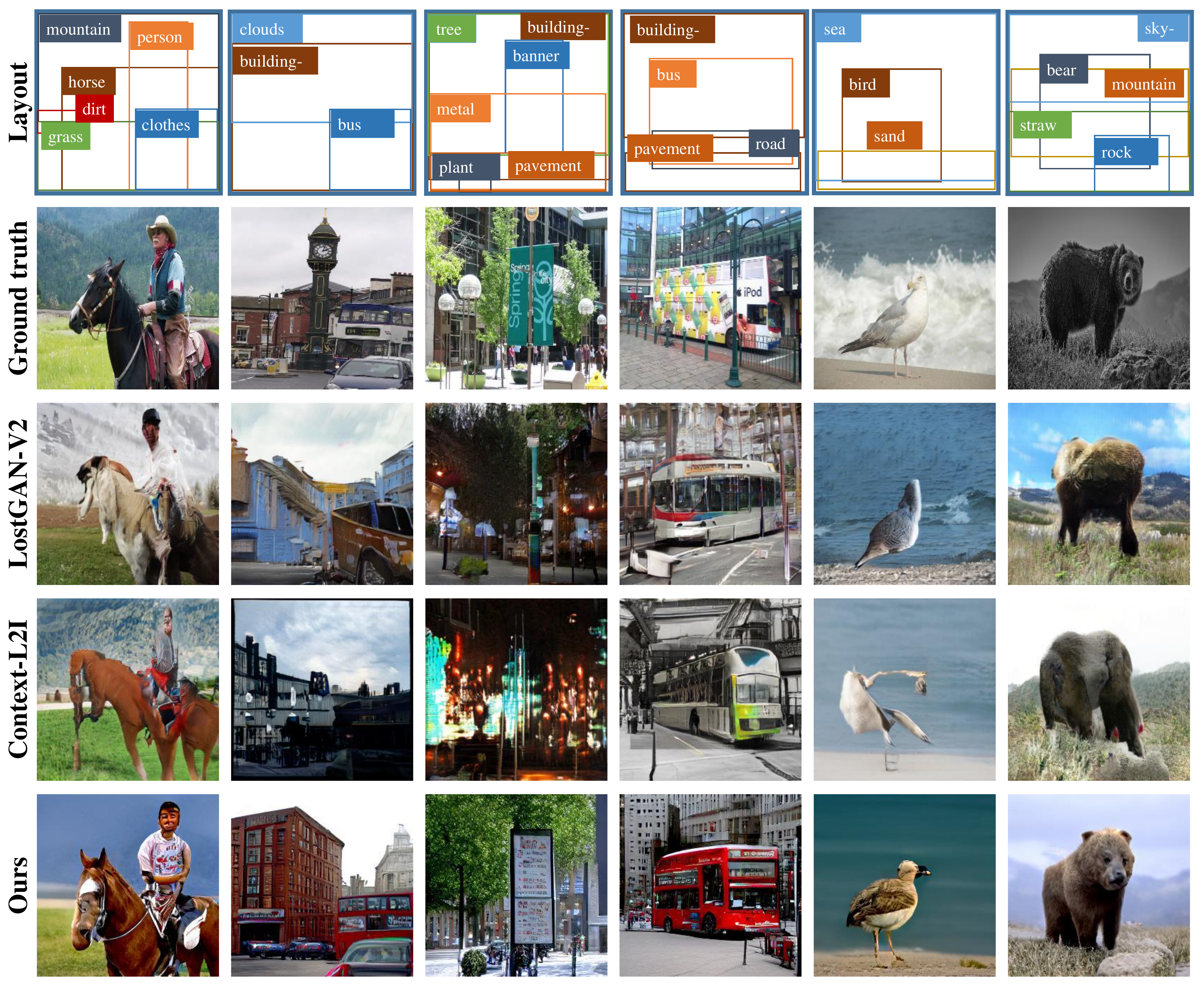}
	\vspace{-3mm}
	\caption{Samples generated from the layouts in COCO-stuff\cite{caesar2018coco} by our method against the most representative baseline model, \ie LostGAN-V2~\cite{sun2021learning} and the state-of-the-art existing model in Table~\ref{tab_SoTA}, \ie Context-L2I~\cite{he2021context}. For all the different scenes, TwFA outperforms the state-of-the-art model with finer instance structures. More results are demonstrated in supplementary materials.
    }\vspace{-2 mm}
	\label{fig:sota}
\end{figure*}

\section{Experiments}

In this section, we first introduced our experimental settings that include training/testing datasets, and evaluation metrics. Then, we carried out quantitative and qualitative comparisons between our method and state-of-the-art CNN-based and transformer-based layout-to-image generation methods. Ablation studies are performed to validate effectiveness of the proposed focal attention. Finally, we performed few-shot complex scene image generation. 

\subsection{Experimental settings}

\noindent \textbf{Datasets.}
Following previous layout to image generation papers, we validate the proposed TwFA and state-of-the-art methods on two datasets: COCO-stuff~\cite{lin2014microsoft,caesar2018coco} and Visual Genome~\cite{krishna2017visual}. COCO-stuff dataset is an expansion
of the Microsoft Common Objects in Context (MS-COCO) dataset, which includes 91 stuff classes and 80 object classes. Visual Genome (VG) is a complex scene understanding dataset that contains annotations such as bounding
boxes, object attributes, relationships, \etc. Following existing works~\cite{sun2021learning,he2021context,li2021image}, we only employed scene images, bounding boxes and labels in both datasets. In this paper, for fair comparisions, all models are trained on the resolution of $256\times256$. 

\noindent \textbf{Implementation Details.} In the initial stage, the shot edge of each training image is first scaled to 296 pixels, keeping the image's aspect ratio unchanged. Then the pre-trained VQ-GAN is utilized to tokenize each $256\times256$ px crop into $16 \times 16$ tokens. The codebook size is set to 8192. In the second stage, we leverage these tokens to train our TwFA (24 layers, 16 attention heads, and 1024 embedding dimensions) and only implement the focal attention on the instance objects. The dropout rate is set to 0.1. When testing, we directly resize the images to $256\times256$ without any other augmentation.

\begin{table*}[t] 
\centering
\begin{tabular}{c|cc|cc|cc|cc} 
\hline
                     & \multicolumn{2}{c|}{FID $\downarrow$}        & \multicolumn{2}{c|}{SceneFID $\downarrow$} & \multicolumn{2}{c|}{Inception Score $\uparrow$}                   & \multicolumn{2}{c}{Diversity Score $\uparrow$}                  \\
                     & COCO           & VG             & COCO          & VG            & COCO                & VG                  & COCO               & VG                 \\ \hline
LostGAN-V2~\cite{sun2021learning}           & 42.55          & 47.62          & 22.00         & 18.27         & 18.01{\footnotesize $\pm$0.50}          & 14.10{\footnotesize $\pm$0.38}          & 0.55{\footnotesize $\pm$0.09}          & 0.53{\footnotesize $\pm$0.09}          \\ \hline
OCGAN~\cite{sylvain2021object}                & 41.65          & 40.85          & -             & -             & -                   & -                   & -                  & -                  \\ \hline
HCSS~\cite{jahn2021high}                 & 33.68          & 19.14          & 13.36         & 8.61          & -                   & -                   & -                  & -                  \\ \hline
LAMA~\cite{li2021image}                 & 31.12          & 31.63          & 18.64         & 13.66         & -                   & -                   & 0.48{\footnotesize $\pm$0.11}          & 0.54{\footnotesize $\pm$0.09}          \\ \hline
Context-L2I*~\cite{he2021context}                 & 29.56          & -             & 14.40         & -               & 18.57{\footnotesize $\pm$0.54}                   & -                   & 0.65{\footnotesize $\pm$0.00}          & -               \\ \hline
Ours                 & \textbf{22.15} & \textbf{17.74} & \textbf{11.99} & \textbf{7.54} & \textbf{24.25{\footnotesize $\pm$1.04}} & \textbf{25.13{\footnotesize $\pm$0.66}} & \textbf{0.67{\footnotesize $\pm$0.00}} & \textbf{0.64{\footnotesize $\pm$0.00}} \\ \hline
\end{tabular}
 \vspace{-2 mm}
\caption{Quantitative results on COCO-stuff\cite{caesar2018coco} and Visual Genome (VG)~\cite{krishna2017visual}. For fair comparisons, all the results are taken from the original papers and based on the resolution of $256\times256$. `-' means the related value is unavailable in their papers. `*' denotes results on samples from trained models with the official implementation.}  \label{tab_SoTA}
 \vspace{-2mm}
\end{table*}

\noindent \textbf{Evaluation Metrics.}
To evaluate the generation performance over all comparison methods and our TwFA, we adopted five metrics to evaluate the visual realism and diversity of generated complex scene images. They are Inception Score (IS)~\cite{salimans2016improved}, Frechet Inception Distance (FID)~\cite{heusel2017gans}, SceneFID~\cite{sylvain2021object}, Diversity Score (DS)~\cite{zhang2018unreasonable}, and YOLO Scores~\cite{li2021image}. Inception Score (IS) is one of the earliest metrics for automatically evaluating the quality of image generative models. For the FID score, it first extracts image features from a pretrained backbone network (\eg Inception V3 trained on ImageNet dataset), then computes the 2-Wasserstein metric between real-world images and generated images. Similarly, SceneFID is proposed for complex scene generation tasks. It computes the Frechet Inception Distance (FID) on the crops of all objects instead of the whole image. Different from measuring the distribution of generated images,  Diversity Score (DS) compares the difference between the generated image and the real image from the same layout. Additionally, YOLO Scores are employed as an evaluation metric to measure the consistency of generated images' layouts to conditions.

\subsection{Comparisons with Existing Methods}

To validate the effectiveness of the proposed TwFA, we compared our model with both CNN-based and Transformer-based complex scene generation methods. Among them, CNN-based methods include LostGAN-V2~\cite{sun2021learning}, OCGAN~\cite{sylvain2021object}, LAMA~\cite{li2021image}, Context-L2I~\cite{he2021context} and the only transformer-based method is HCSS~\cite{jahn2021high}. For a fair comparison, we adopt their official released pre-trained models or the official reported scores in their papers.

The quantitative results by the involved competitors on both the COCO-stuff and Visual Genome datasets are reported in Table~\ref{tab_SoTA}. 
Among existing methods, Context-L2I~\cite{he2021context} achieved the best overall performance. HCSS~\cite{jahn2021high} employed a transformer with self-attention to perform the complex scene composition modeling task. Since we employed the same texture tokenization strategy, the generation performance depends on how well the transformer can model the composition of complex scenes. Compared to them, ours achieved significant improvement on all metrics. 

In Figure~\ref{fig:sota}, we provide several visual comparisons of the complex scene images generated by different methods based on the same layout. According to the visual comparison, we can observe that pervious methods are capable of generating reasonable texture and patches. According to the generated texture, we can roughly understand the synthesized scene. However, they failed to accurately model the instance structures. For example, the bear generated by LostGAN looks like a collection of the animal fur. According to the visual examples, the proposed TwFA performs better on constructing relationships between objects and modeling structure of instances. More examples can be found in our supplementary material.

\begin{table}[]
\centering
\resizebox{0.46\textwidth}{!}{
\begin{tabular}{c|c|c|c|c}
\hline
         & Grid2      & Grid4      & Grid16     & Ours                \\ \hline
FID $\downarrow$      & 27.64      & 29.01      & 27.04      & \textbf{22.15}      \\ \hline
S-FID $\downarrow$ & 17.26      & 19.43      & 16.80      & \textbf{11.99}      \\ \hline
IS $\uparrow$         & 21.84{\footnotesize $\pm$0.88} & 21.50{\footnotesize $\pm$0.59} & 22.73{\footnotesize $\pm$0.65} & \textbf{24.25{\footnotesize $\pm$1.04}} \\ \hline
\end{tabular}}
 \vspace{-2 mm}
\caption{Comparison of different attention configurations. Grid16 is equivalent to vanilla attention. S-FID denotes SceneFID.}
\label{tab_ablation}
\end{table}

Overall, the superiority of the proposed TwFA is validated on both quantitative metrics and qualitative visual comparison. The metrics such as FID, sceneFID and IS demonstrated the distribution of TwFA-generated images are statistical better then other methods. And TwFA largely improved the visually quality of complex scene generation.  

\begin{table}[t]
\centering
\resizebox{0.45\textwidth}{!}{
\begin{tabular}{c|c|c|c|c|c}
\hline
\begin{tabular}[c]{@{}c@{}}YOLO\\ Scores\end{tabular} & Grid16 & \begin{tabular}[c]{@{}c@{}}Ours\\ w/o oo\end{tabular} & \begin{tabular}[c]{@{}c@{}}Ours\\ w/o op\end{tabular} & \begin{tabular}[c]{@{}c@{}}Ours\\ w/o pp\end{tabular} & \begin{tabular}[c]{@{}c@{}}Ours\\ Full\end{tabular} \\ \hline
$AP_{50}$ $\uparrow$                                                  & 25.97\%  & 25.01\%                                                 & 27.59\%                                                 & 26.00\%                                                 & \textbf{28.20\%}                                               \\ \hline
$AP_{75}$ $\uparrow$                                                  & 17.45\%  & 17.38\%                                                 & 19.00\%                                                 & 17.93\%                                                 & \textbf{20.12\%}                                               \\ \hline
\end{tabular}}
 \vspace{-2 mm}
\caption{Comparison of different interaction configurations in the Connectivity Matrix. ``oo", ``op", and ``pp" denote the object-object, object-patch, and patch-patch interaction respectively.
}
\label{tab_yolo}
\end{table}

\subsection{Ablation Study}

Here, we aim to explore how the attention mechanisms will influence transformer modeling of the complex scene generation. 
Besides the proposed focal attention, we performed other three attention configurations. First, we employed the global self-attention with causal mask that is widely used in language and image generation tasks. When predicting the current token, it performs self-attention with all given tokens, \ie the layout conditions and all patch tokens before it. Inspired by the recent popular local attentions proposed in vision transformer, we test additional two attention mechanisms with sliding window size 2 and 4, \ie `Grid2' and `Grid4'. Specifically, when predicting the current patch token, besides layout conditions, the model only attended on the given patch tokens within the 2D sliding windows. In our model, since the size of compressed token map is $16\times16$, therefore the setting of global self-attention is equivalent to `Grid16'. 

\begin{figure}[!t]
	\centering \vspace{-5 mm}
	\includegraphics[width=\linewidth]{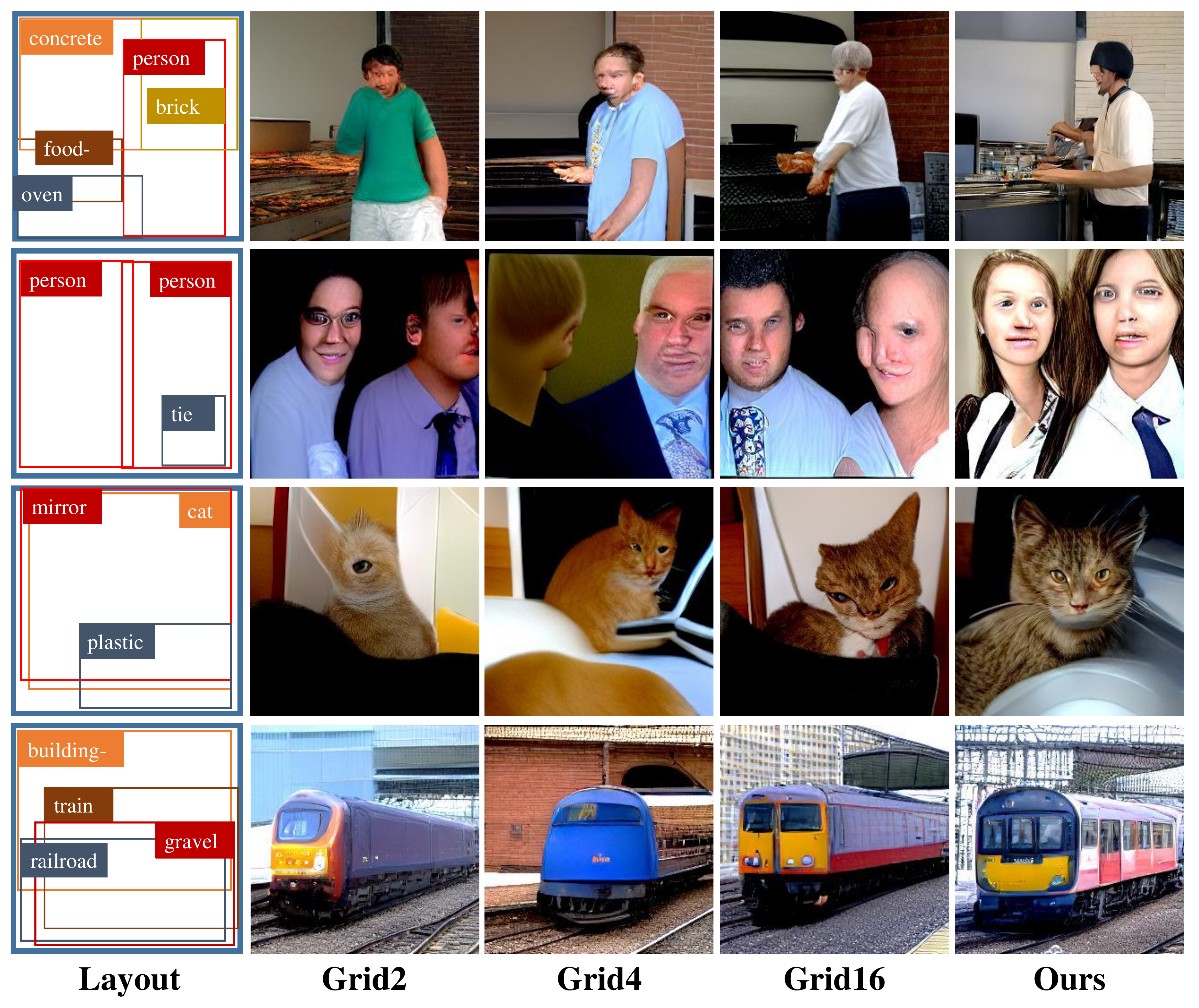}
	\caption{Qualitative ablative results of different attention configurations. Our focal attention achieves the best results.}
	\label{fig:ablation}
\end{figure}

The quantitative results are reported in Table~\ref{tab_ablation}. We can see that our focal attention achieved the best performance on all metrics. Interestingly, we find that both Grid16 and Grid2 perform better than Grid4. It means either modeling global dependencies or local dependencies would contribute to increasing the performance, yet selecting a window size larger than 2 may decrease the performance. A possible explanation is that the relatively large window size is easier to break down the dependencies within and outside the bounding box, and meanwhile, fail to model global relationships. Compared with them, the proposed focal attention better utilizes the information provided by layouts. The same conclusion can be drived from Figure~\ref{fig:ablation}. As illustrated in the second row, the models with a larger window size failed to generate another face correctly.

Additionally, we ablate different connectivity matrix components, \textit{i.e.}, object-object, object-patch, and patch-patch interaction, to investigate their effect to object composition modeling in complex scene generation.
Here, we utilize YOLO Scores to evaluate the alignment and fidelity of generated objects.
As shown in Table~\ref{tab_yolo}, we have the following observations:
\textbf{1)} The lack of the object-object interaction leads to a huge decrease. It indicates the global context greatly influences the object structure reasoning;
\textbf{2)} Both ``Ours w/o op'' and ``Ours w/o pp'' performs better than Grid16, which suggests the well-designed object-patch/patch-patch interaction is essential to generating an photorealistic object;
\textbf{3)} By integrating all the interactions, the TwFA improves Grid16 from 17.45\% to 20.12\% (\textit{+15.3\%}) and from 25.97\% to 28.20\% (\textit{+8.6\%}) on $AP_{75}$ and $AP_{50}$, respectively.

\subsection{Few-shot Complex Scene Generation}

As aforementioned, benefiting from the accurate relationship modeling, the well-trained TwFA has the potential to perform few-shot complex scene image generation. Specifically, through fine-tuning on a few of images that contain unseen objects, we hope our model can be trained to generate this kind of objects giving new layouts. 

For providing both quantitative and qualitative analysis, we trained a baseline (\ie Grid16*) and a Ours* using COCO-stuff training data that removed all zebra images. Thus, zebra images in the original COCO-stuff dataset can act as the additional images for few-shot learning. Here, we choose zebra images for two reasons, (i) zebra is a kind of challenging target to synthesize, since it has a unique texture and complex structure/pose; (ii) there are relatively large amount of zebra images ($\sim$1500) in COCO-stuff datasets, which contributes to more accurate quantitative measurement (\eg FID prefer testing on more images). 

\begin{figure}[!t]
	\centering \vspace{-5 mm}
	\includegraphics[width=\linewidth]{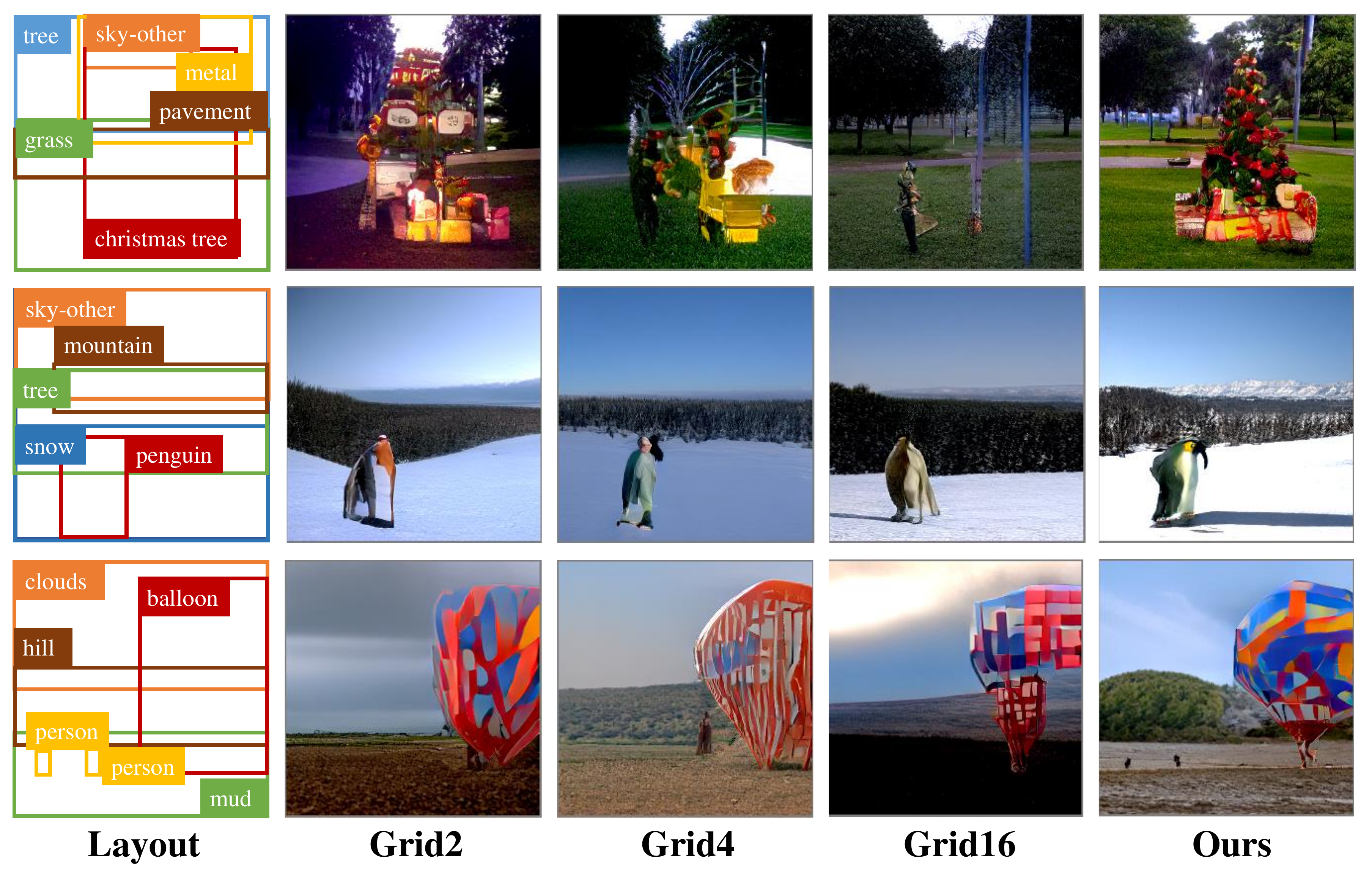} \vspace{-5 mm}
	\caption{Examples of few-shot results. 
    The novel classes are the Christmas tree, penguin, and hot air balloon.
	TwFA outperforms all the baseline model with finer structures and details.
    }
	\label{fig:fewshot}
\end{figure}
\begin{table}[t]
\centering
\begin{tabular}{c|cc|cc}
\hline
             & \multicolumn{2}{c|}{Grid16*}          & \multicolumn{2}{c}{Ours*}                             \\ \hline
\# of Shot       & \multicolumn{1}{c|}{FID $\downarrow$} & Obj-FID $\downarrow$ & \multicolumn{1}{c|}{FID $\downarrow$} & Obj-FID $\downarrow$ \\ \hline
20           & \multicolumn{1}{c|}{39.47} & 35.62   & \multicolumn{1}{c|}{\textbf{36.34}} & \textbf{31.17} \\ \hline
30           & \multicolumn{1}{c|}{39.32} & 34.59   & \multicolumn{1}{c|}{\textbf{34.87}} & \textbf{29.33} \\ \hline
40           & \multicolumn{1}{c|}{37.36} & 31.34   & \multicolumn{1}{c|}{\textbf{34.30}} & \textbf{28.87} \\ \hline
50           & \multicolumn{1}{c|}{37.47} & 32.81   & \multicolumn{1}{c|}{\textbf{31.53}} & \textbf{26.73} \\ \hline \hline
Full trained & \multicolumn{1}{c|}{30.28} & 21.96   & \multicolumn{1}{c|}{\textbf{24.33}}          & \textbf{21.66}          \\ \hline
\end{tabular}  \vspace{-2 mm}
\caption{Qualitative few shot results. Obj-FID only computes the FID score on the crops of the novel class with a size of $224 \times 224$.}
\label{tab_fewshot} \vspace{-3 mm}
\end{table}

As shown in Table~\ref{tab_fewshot}, all experiments on 20/30/40/50 shots show the superiority of our Ours*. Meanwhile, accompany with the increasing of shot number, better generation performance can be achieved. In Figure~\ref{fig:fewshot}, three new classes (Christmas tree, penguin, and hot air balloon) with 2 samples are employed to fine-tune both the baseline and our TwFA. From the visual examples, we can see our TwFA show better instance structure compared to the baseline model. Since space limitation, more few-shot generation results are reported in the supplementary material.


\section{Conclusion}

In this paper, we presented a novel Transformer with Focal Attention (TwFA) to disentangle the modeling between object-level relationships and patch-level instance structures, and introduce the composition prior from spatial layouts into image compositions.
Compared with CNN-based and Transformer-based methods, TwFA enables the model to capture the inherent instance structures, and increase the data efficiency to alleviate the burden on few-shot learning with limited data.
With extensive experiments and visualizations on both COCO-stuff and Visual Genome datasets, the proposed TwFA demonstrates its superiority over the SoTA methods on both L2I and few-shot L2I tasks.




{\small
\bibliographystyle{ieee_fullname}
\bibliography{main.bbl}
}

\clearpage
\section{Appendix}
In this appendix, we first demonstrate the model's reconfigurable and diverse generation ability in Section~\ref{sec:6.1}. Next, we perform visual comparison with more SoTA methods in Section~\ref{sec:6.2}. Then, we provide more L2I Examples in Section~\ref{sec:6.3} and more Few-shot L2I Examples in Section~\ref{sec:6.4}. Finally, the limitations and broader impacts will be discussed in Section~\ref{sec:6.5} 


\begin{figure}[!t]
	\centering
	\includegraphics[width=\linewidth]{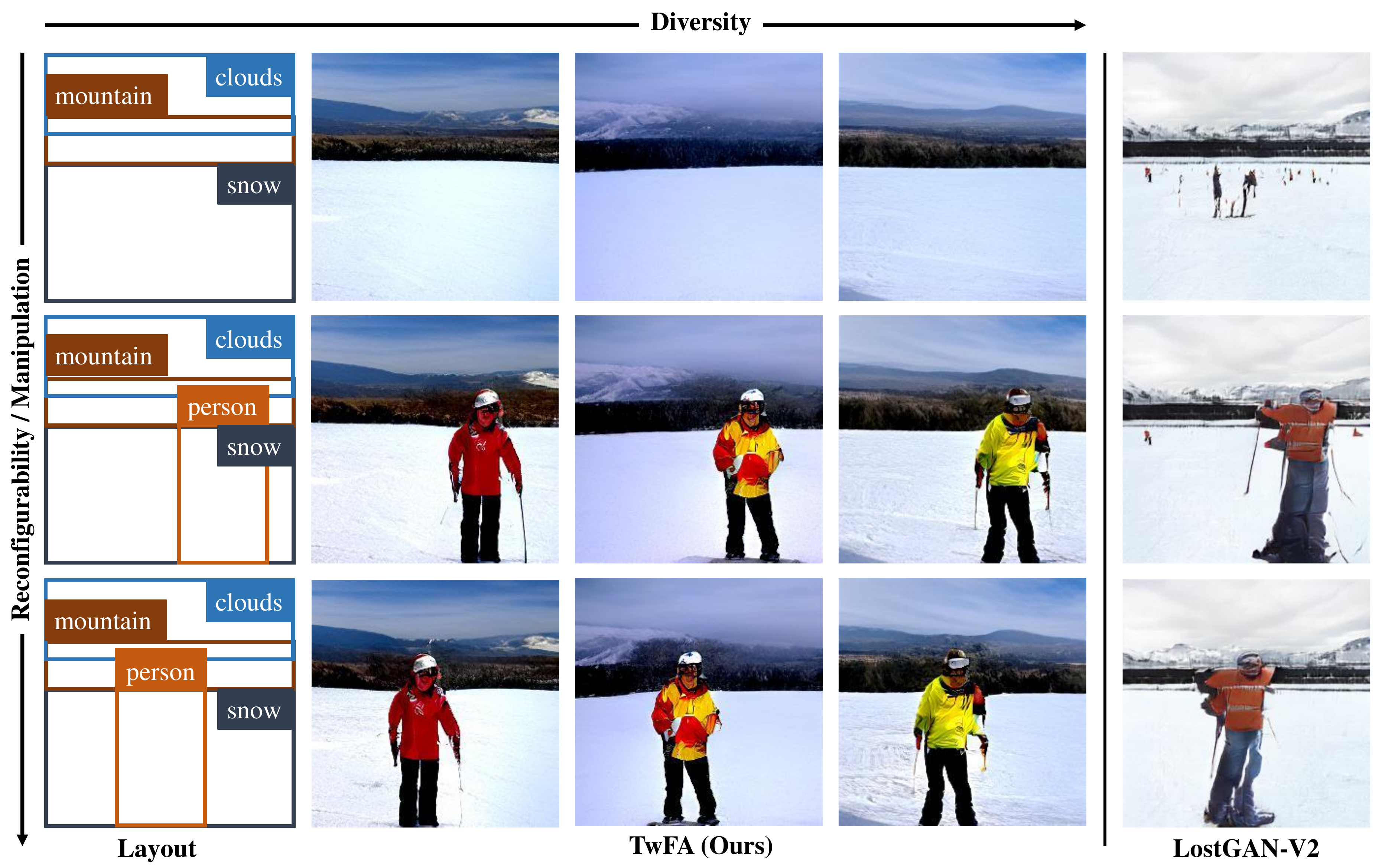}
	\caption{Examples of reconfigurable and diverse results (Section~\ref{sec:6.1}). 
	Each row shows diverse generated images from the same layout on the left. Each column shows effects of the reconfiguration/manipulation by adding or moving objects. Compared with LostGANs-V2, TwFA synthesizes cleaner snow and better human structure. 
    }
	\label{app_fig:reconfigurable}
\end{figure}

\subsection{Reconfigurable and Diverse Generation}
\label{sec:6.1}

\noindent \textbf{Reconfigurability.}
TwFA is reconfigurable and easy to manipulate as shown in each column of Figure~\ref{app_fig:reconfigurable}.
We can find that TwFA successfully maintains the styles after manipulating the layouts, \textit{e.g.}, adding or moving objects.
The reason is that focal attention insulates the patch-level interaction, \textit{i.e.}, each patch only attends on related patches inner the same object and will not be affected by other unrelated patches.

\noindent \textbf{Diversity.}
TwFA generates diverse images by multinomial resampling strategy, \textit{i.e.}, we randomly sample each patch token according to its probability distribution $p(s_i)$ in Eq.(2).
This multinomial resampling strategy introduces uncertainty into the generation process, leading to generation diversity. As shown in each row of Figure~\ref{app_fig:reconfigurable}, TwFA can generate clothes of various colors and mountains of different styles. 




\subsection{Visual Comparison with More SoTAs}
\label{sec:6.2}

\begin{figure*}[!t]
	\centering
	\includegraphics[width=\linewidth]{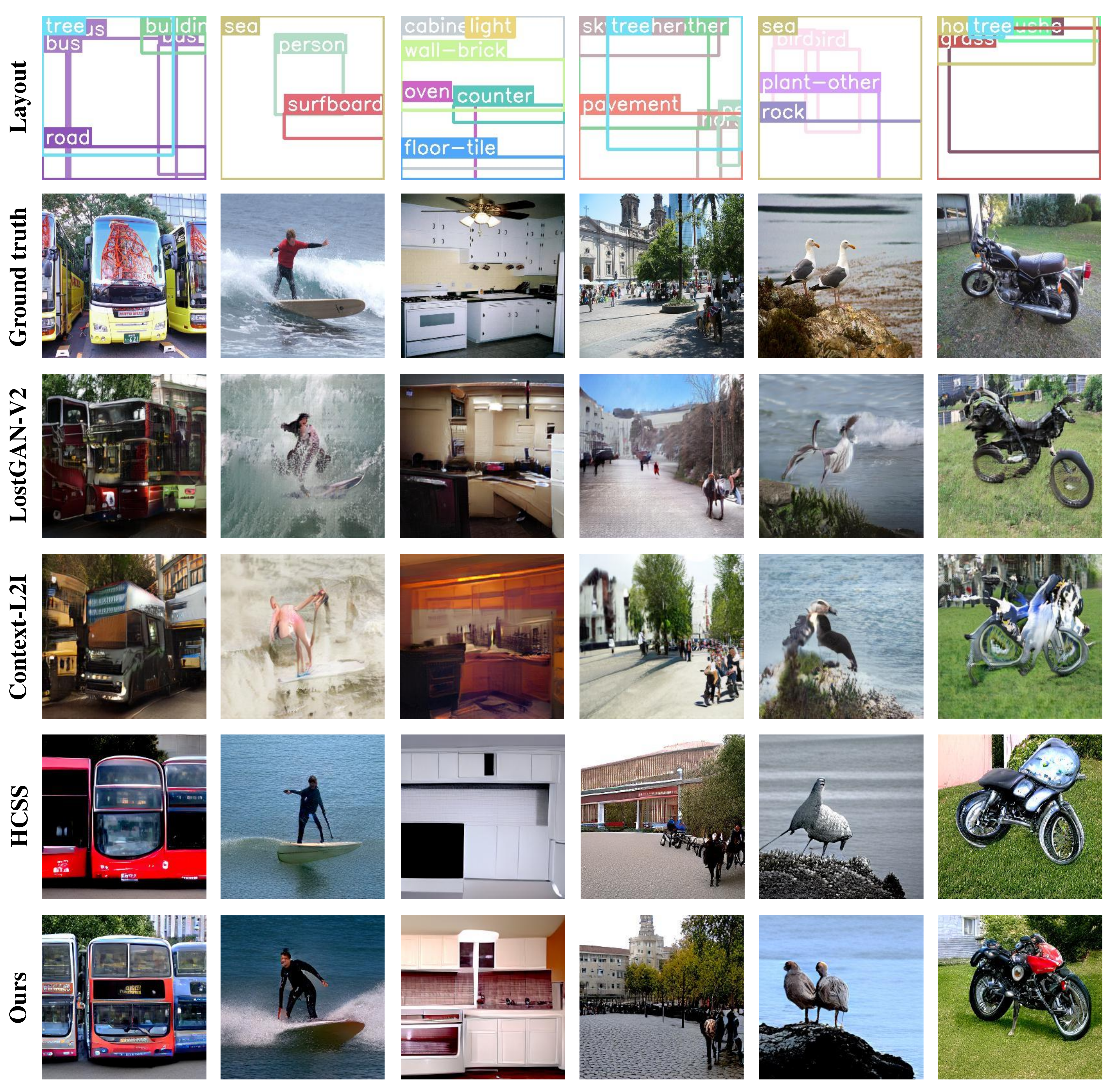}
	\vspace{-5mm}
	\caption{Comparisons with SoTAs (Section~\ref{sec:6.2}). Our method is compared against the most representative baseline model LostGAN-V2~\cite{sun2021learning}, the existing state-of-the-art model Context-L2I~\cite{he2021context}, and the transformer-based method HCSS~\cite{jahn2021high}. For all different scenes, TwFA outperforms the state-of-the-art model with more reasonable relationships, finer instance structures, and textures.
    }
	\label{app_fig:sota}
\end{figure*}

\begin{figure*}[!t]
	\centering
	\includegraphics[width=\linewidth]{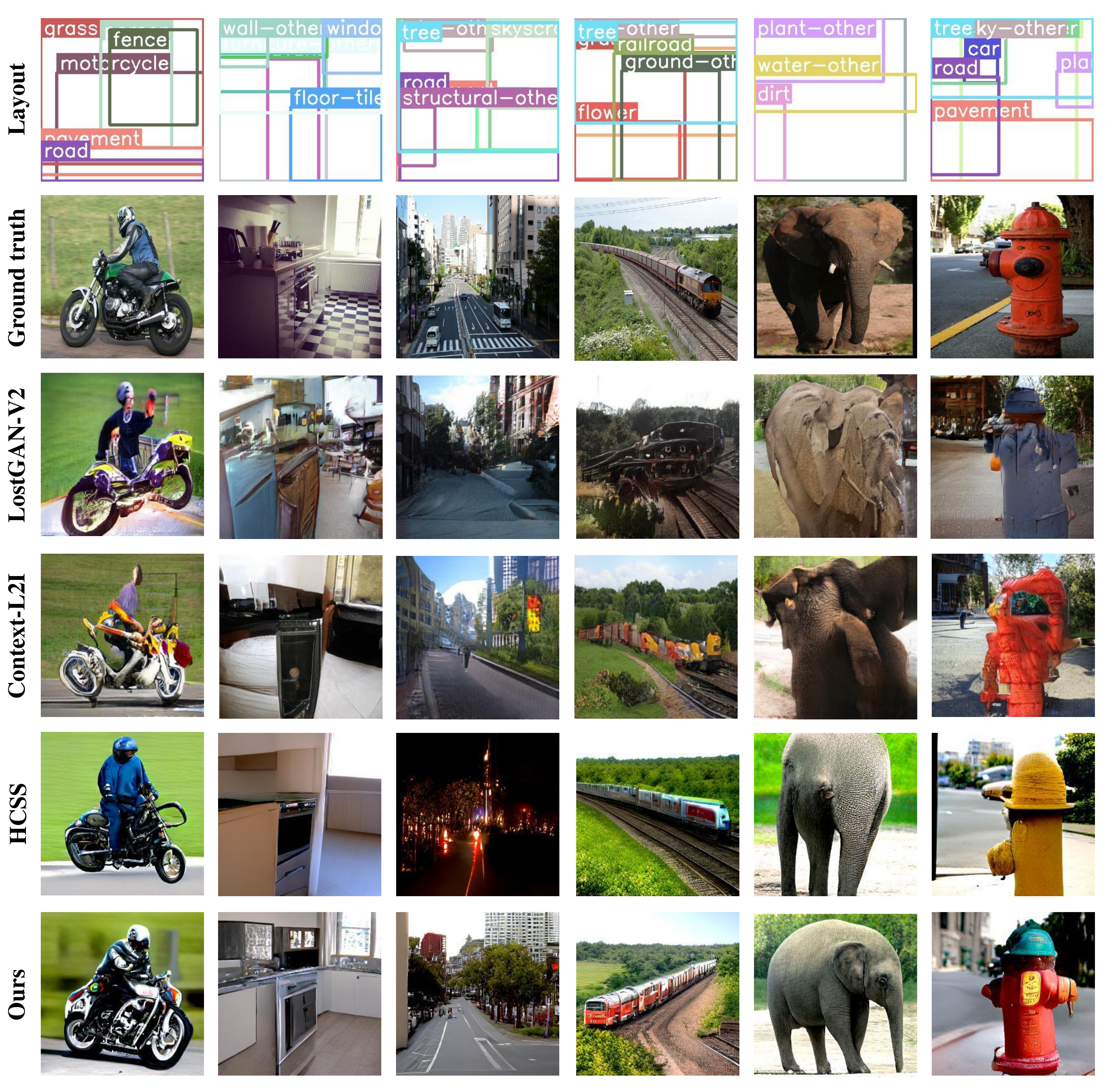}
	\vspace{-5mm}
	\caption{Comparisons with SoTAs (Section~\ref{sec:6.2}). Our method is compared against the most representative baseline model LostGAN-V2~\cite{sun2021learning}, the existing state-of-the-art model Context-L2I~\cite{he2021context}, and the transformer-based method HCSS~\cite{jahn2021high}. For all different scenes, TwFA outperforms the state-of-the-art model with more reasonable relationships, finer instance structures, and textures.
    }
	\label{app_fig:sota2}
\end{figure*}

We perform visual comparison with more SoTA methods, including CNN-based (LostGAN-V2~\cite{sun2021learning}, Context-L2I~\cite{he2021context}) and Transformer-based (HCSS~\cite{jahn2021high}), in Figure~\ref{app_fig:sota} and Figure~\ref{app_fig:sota2}.
Compared with existing SoTA methods, the proposed TwFA can synthesize 
\textbf{1)} More reasonable object-level relationships, \textit{e.g.}, person-surfboard, bird-bird, and person-motorcycle;
\textbf{2)} Clearer patch-level instance structures, \textit{e.g.}, oven, bird, and motorcycle;
\textbf{3)} Refiner pixel-level textures, \textit{e.g.}, rock, grass, and pavement.


\subsection{More L2I Examples}
\label{sec:6.3}
\begin{figure*}[!t]
	\centering
	\includegraphics[width=0.96\linewidth]{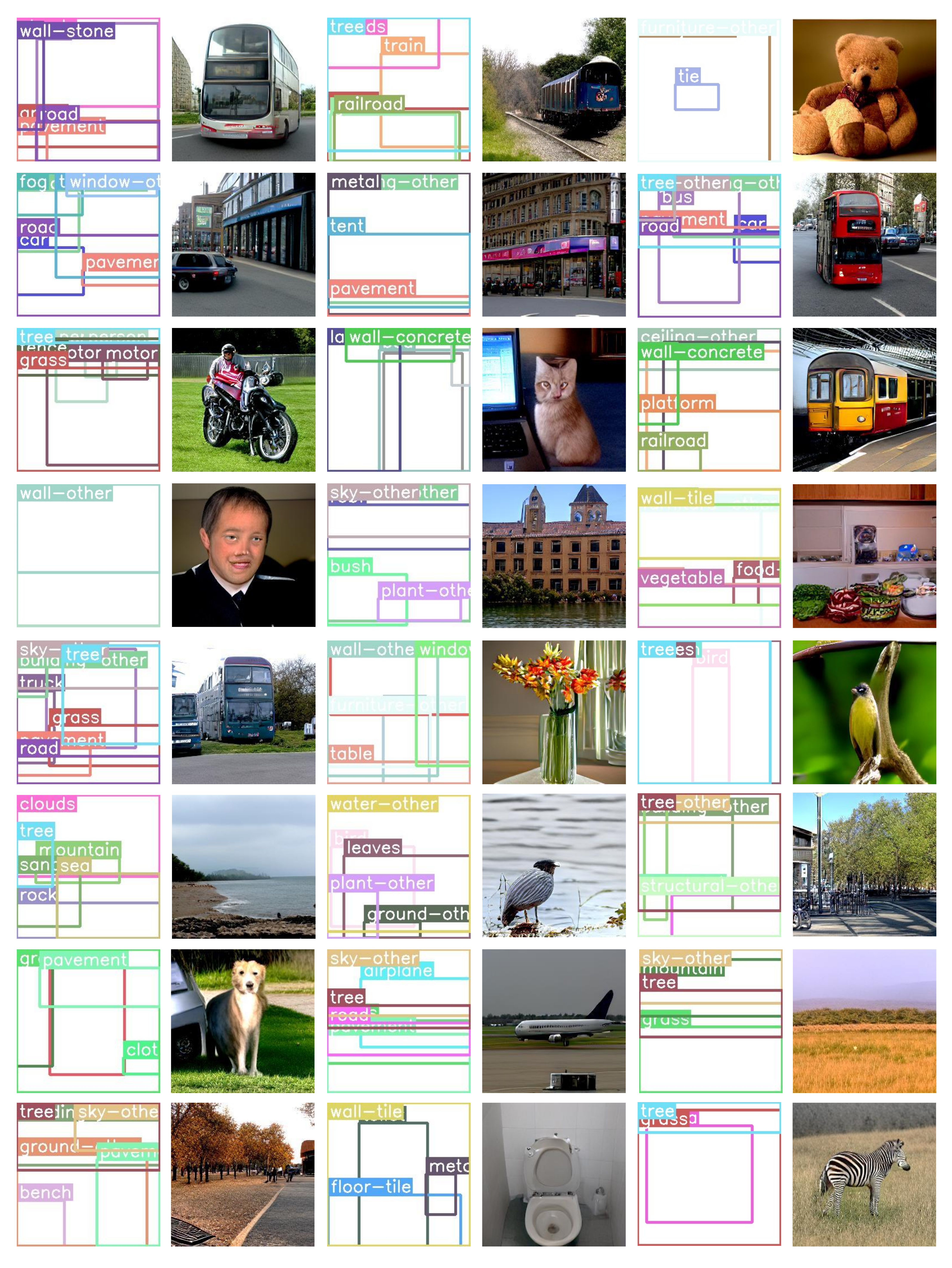}
	\vspace{-4mm}
	\caption{More L2I examples (Section~\ref{sec:6.3}). All images are generated by the proposed TwFA according to the given layout on the left.
    }
	\label{app_fig:result1}
\end{figure*}

\begin{figure*}[!t]
	\centering
	\includegraphics[width=0.96\linewidth]{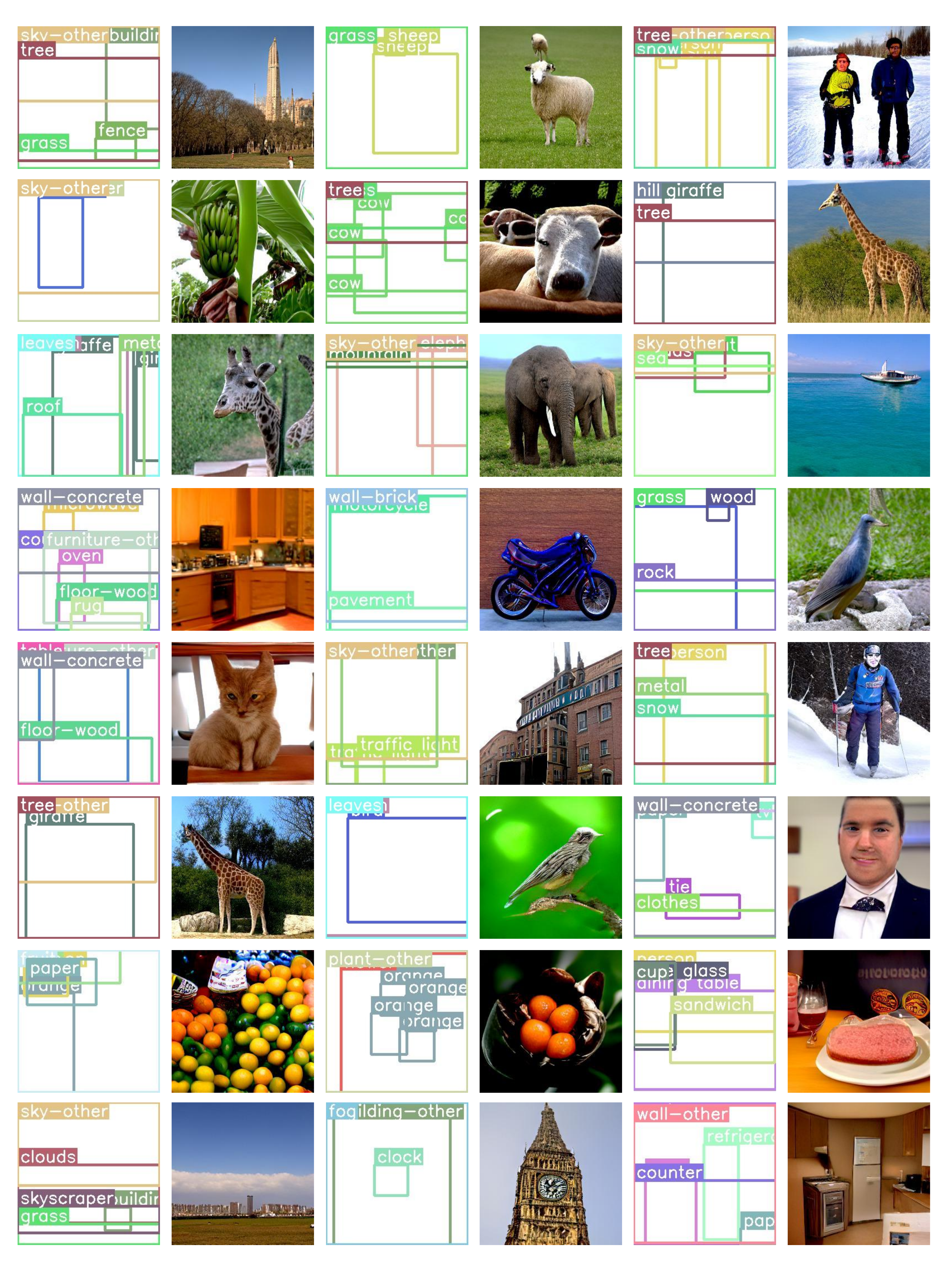}
	\vspace{-4mm}
	\caption{More L2I examples (Section~\ref{sec:6.3}). All images are generated by the proposed TwFA according to the given layout on the left.
    }
	\label{app_fig:result2}
\end{figure*}

\begin{figure*}[!t]
	\centering
	\includegraphics[width=0.96\linewidth]{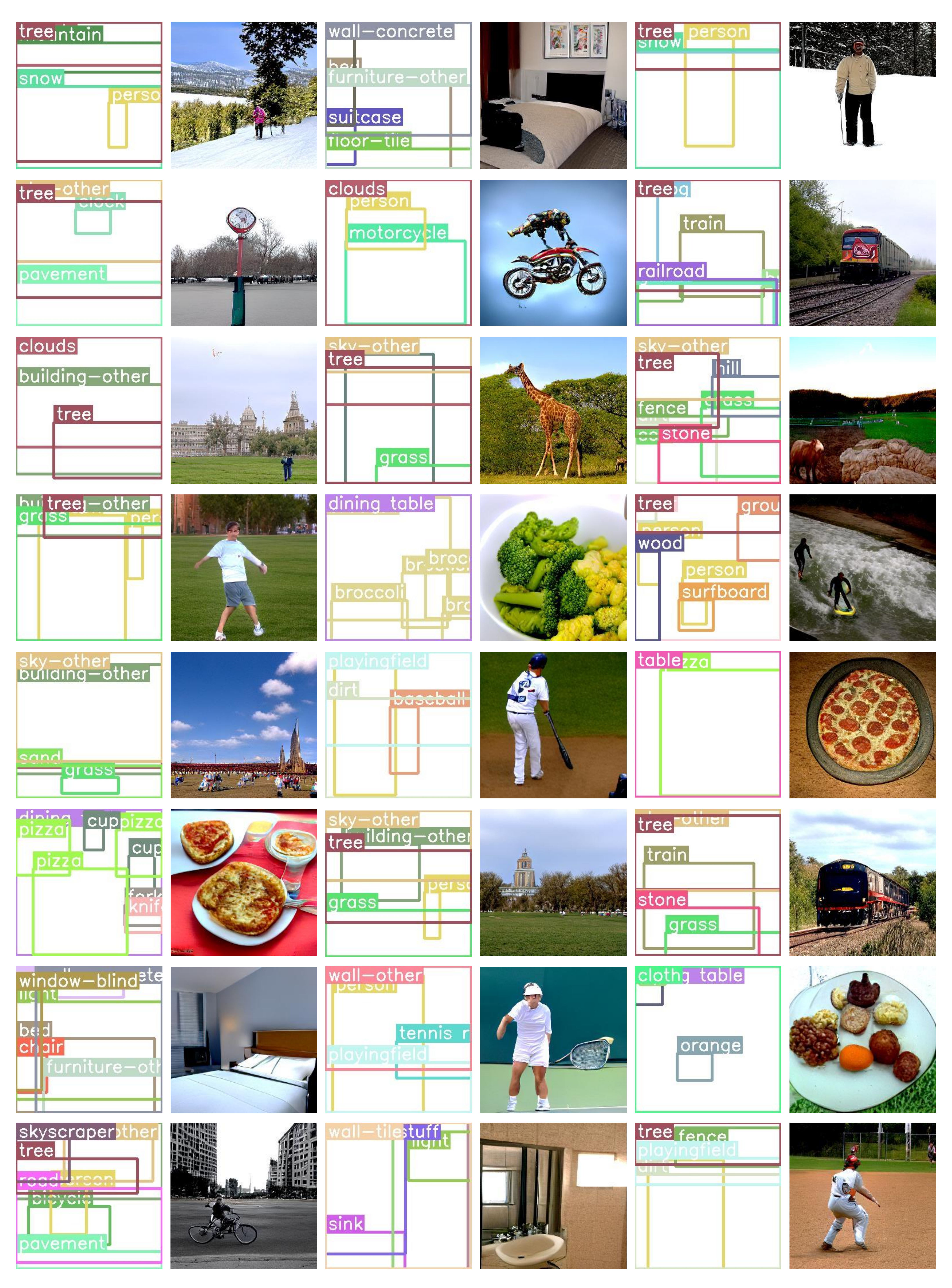}
	\vspace{-4mm}
	\caption{More L2I examples (Section~\ref{sec:6.3}). All images are generated by the proposed TwFA according to the given layout on the left.
    }
	\label{app_fig:result3}
\end{figure*}

\begin{figure*}[!t]
	\centering
	\includegraphics[width=0.96\linewidth]{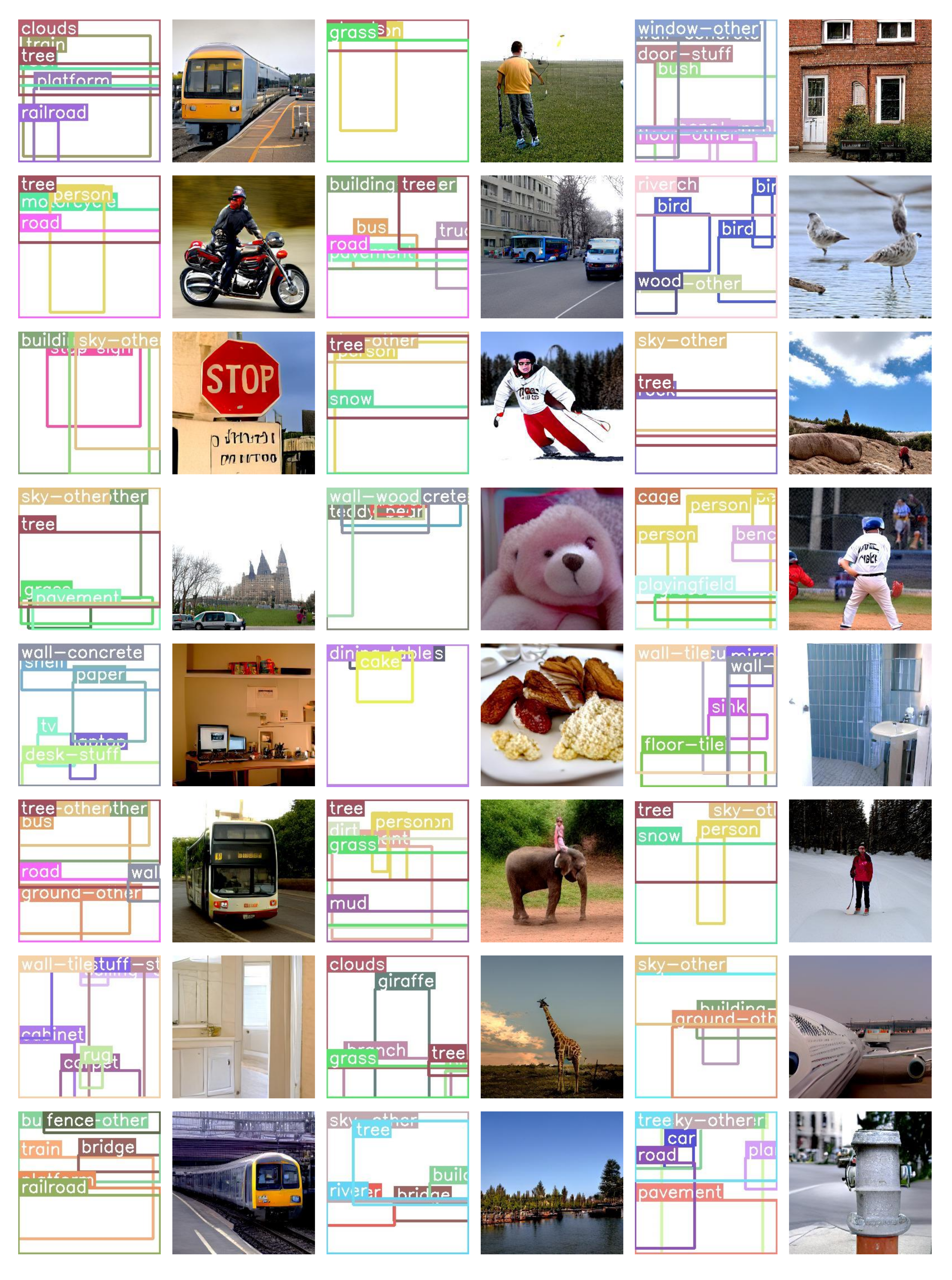}
	\vspace{-4mm}
	\caption{More L2I examples (Section~\ref{sec:6.3}). All images are generated by the proposed TwFA according to the given layout on the left.
    }
	\label{app_fig:result4}
\end{figure*}

\begin{figure*}[!t]
	\centering
	\includegraphics[width=0.96\linewidth]{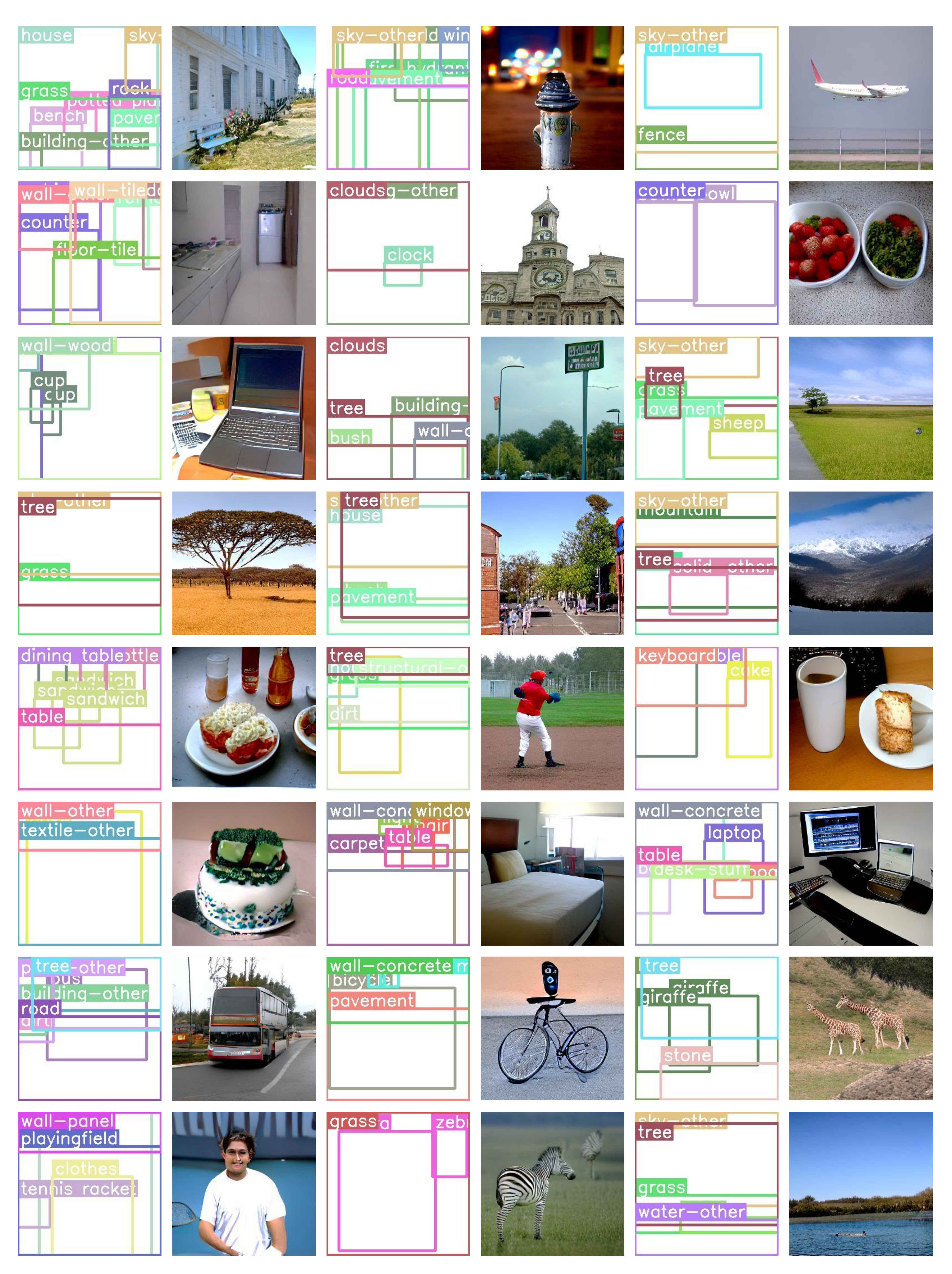}
	\vspace{-4mm}
	\caption{More L2I examples (Section~\ref{sec:6.3}). All images are generated by the proposed TwFA according to the given layout on the left.
    }
	\label{app_fig:result5}
\end{figure*}

More samples generated by TwFA are shown in Figure~\ref{app_fig:result1} -- Figure~\ref{app_fig:result5}.
According to these results, several advantages of TwFA are observed: 
\textbf{1)} As illustrated in Figure~\ref{app_fig:result1}, the reflection on the bus windows/river makes the images to look more realistic. \textbf{2)} The different textures of fur contribute to the fidelity for different animals, such as the cow, cat, and bird, shown in Figure~\ref{app_fig:result2}. \textbf{3)} The well-generated structure for an object also reduces the unreality, (\textit{i.e.}, bus, giraffe, and train). \textbf{4)} The shadow (\textit{i.e.}, on the ground) conforms to the laws of physics, such as the last image in the sixth row of Figure~\ref{app_fig:result1}. All of these advances make the generated images more realistic.

\subsection{More Few-shot L2I Examples}
\label{sec:6.4}
\begin{figure}[!t]
	\centering
	\includegraphics[width=\linewidth]{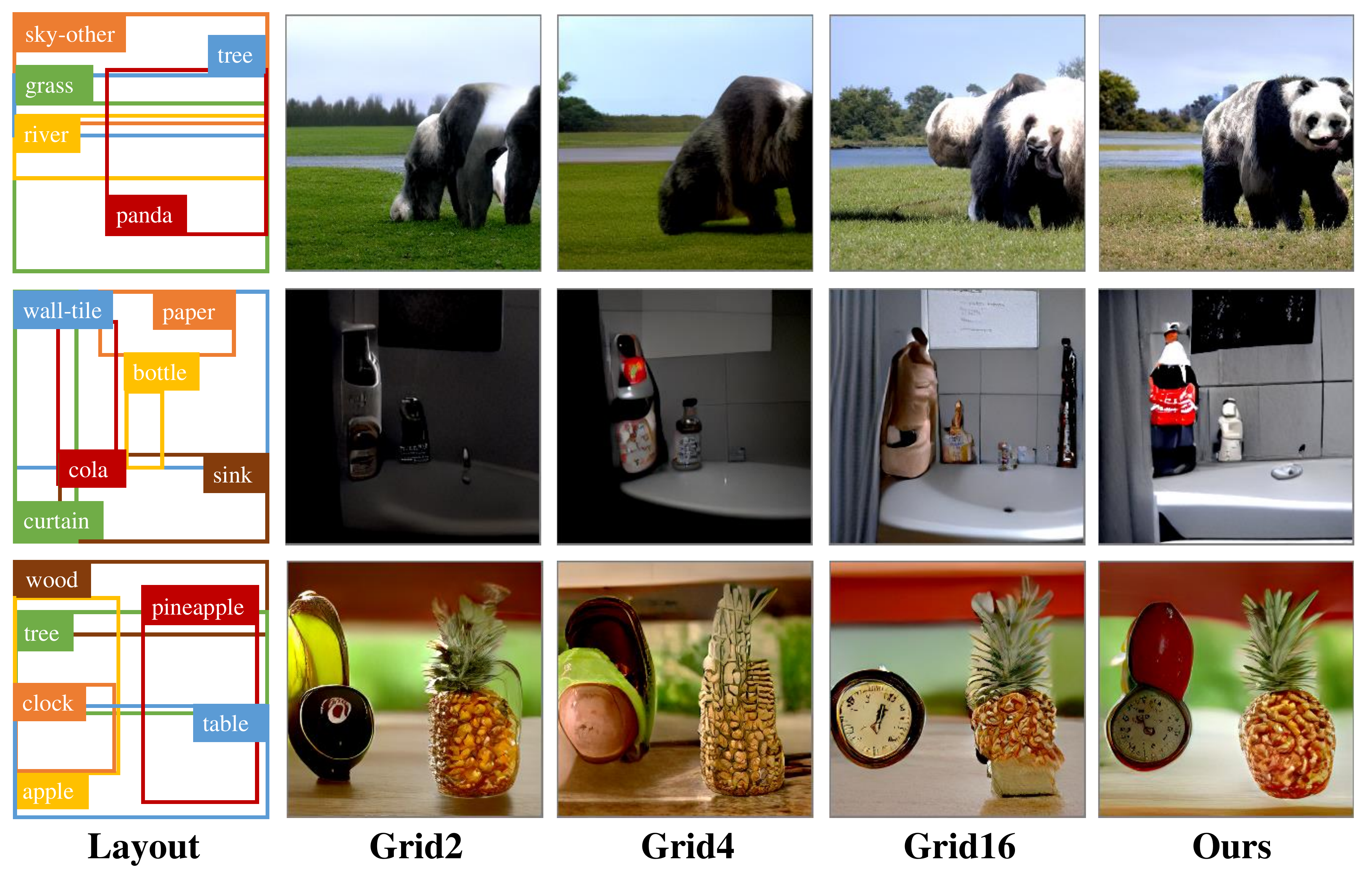}
	\caption{Examples of few-shot results (Section~\ref{sec:6.4}). The novel classes from the first row to the last one are the panda, cola, and pineapple, whose positions are annotated with red rectangles in layouts. TwFA outperforms all the baseline model with finer structures and details.
    }
	\label{app_fig:fewshot}
\end{figure}

To further demonstrate the advance of our method in few-shot complex scene generation, we conducted more experiments with different novel categories. The results are illustrated in Figure~\ref{app_fig:fewshot}.

\noindent \textbf{Settings.}
Few-shot complex scene generation consists of two steps:
\textbf{1)} The first step is to prepare training data with annotated novel classes. To simplify the image annotation process, we randomly select several segmented novel objects and attach them into the images of COCO-stuff. Therefore, we can just annotate the bounding box positions of the novel classes. The rest annotations of the attached images derive from the dataset. \textbf{2)} The second step is to execute the few-shot implementation based on the models trained on the full COCO-stuff~\cite{caesar2018coco} dataset.
To show the extreme situation, all the experiments are conducted in a setting of 2-shot. The novel classes include balloon, panda, penguin, Christmas tree, cola, and pineapple.

\noindent \textbf{Comparisons.}
In this section, we discuss the few-shot result comparisons between our TwFA and other baseline models: \textbf{1)} As illustrated in Figure~\ref{app_fig:fewshot}, our method can learn better instance structures, such as the balloon, panda, and pineapple. As for the penguin, the beak is generated more vividly than other baseline models. \textbf{2)} Our model not only learns how to synthesize the structure of an object, but also learns the interaction with the surroundings for the novel objects (\textit{i.e.}, the penguin's shadow on the snow). \textbf{3)} Our model can reduce the interference of other stuff objects. 
In detail, Grid16 fails to synthesize the Christmas tree with the background of trees. Meanwhile, when generating the cola bottle, the disturbance from the surrounding patches leads to generating an arbitrary bottle instead of a cola bottle. 
In conclusion, TwFA outperforms all the baseline models with finer structures and details.

\subsection{Limitations and Broader Impacts.}
\label{sec:6.5}

\noindent \textbf{Limitations.}
Though the proposed TwFA achieves impressive performance on (few-shot) L2I tasks, it relies on the composition prior from given layouts which may not be available in other complex scene generation tasks, \textit{e.g.}, textual descriptions~\cite{hinz2020semantic} and scene graphs~\cite{johnson2018image}. How to extend our method to those tasks remains open. We leave this as a promising direction to explore in the future.


\noindent \textbf{Broader impacts.}
The proposed method synthesizes images based on learned patterns of the training dataset and as such will reflect biases in those data, including ones with negative societal impacts. The model may generate inexistent images with unexpected content. These issues warrant further research and consideration when generating images based on this work.



\end{document}